\documentclass[runningheads]{llncs}
\usepackage{multirow,array}

\usepackage{graphicx}
\usepackage{amsmath, ,amssymb} % define this before the line numbering.

\usepackage{color}
\usepackage{subfigure}
\usepackage[width=122mm,left=12mm,paperwidth=146mm,height=193mm,top=12mm,paperheight=217mm]{geometry}

\begin{document}

\pagestyle{headings}
\mainmatter

\title{Interactive Medical Image Segmentation via Point-Based Interaction and Sequential Patch Learning} % Replace with your title

\titlerunning{A very long title}
\authorrunning{authors running}

\author{Jinquan Sun$^1$, Yinghuan Shi$^1$, Yang Gao$^1$, Lei Wang$^2$, Luping Zhou$^2$, Wanqi Yang$^3$, Dinggang Shen$^4$}

%Please write out author names in full in the paper, i.e. full given and family names. 
%If any authors have names that can be parsed into FirstName LastName in multiple ways, please include the correct parsing, in a comment to the volume editors:
%\index{Lastnames, Firstnames}
%(Do not uncomment it, because you may introduce extra index items if you do that...)

%Department,\\
%	University\\
%	\email{ \{author1,author2\}@univ.edu}   

\institute{
	$^1$Nanjing University\\
	\email{\{jasonjinquan\}@gmail.com, \{syh, gaoy\}@nju.edu.cn}\\
	$^2$University of Wollongong\\
	\email{\{leiw, lupingz\}@uow.edu.au}
	$^3$Nanjing Normal University\\
	\email{\{yangwq\}@njnu.edu.cn}\\
	$^4$University of North Carolina at Chapel Hill\\
	\email{\{dgshen\}@med.unc.edu}\\
}

\maketitle

\begin{abstract}
Due to low tissue contrast, irregular object appearance, and unpredictable location variation, segmenting the objects from different medical imaging modalities (\emph{e.g.}, CT, MR) is  considered as an important yet challenging task. In this paper, we present a novel method for interactive medical image segmentation with the following merits. (1) Our design is fundamentally different from previous pure patch-based and image-based segmentation methods. We observe that during delineation, the physician repeatedly check the inside-outside intensity changing to determine the boundary, which indicates that \emph{comparison in an inside-outside manner is extremely important}.  Thus, we innovatively model our segmentation task as learning the representation of the bi-directional sequential patches, starting from (or ending in) the given central point of the object. This can be realized by our proposed ConvRNN network embedded with a gated memory propagation unit.
 (2) Unlike previous interactive methods (requiring bounding box or seed points), we only ask the physician to merely click on the rough central point of the object before segmentation, which could simultaneously enhance the performance and reduce the segmentation time. 
(3) We utilize our method in a multi-level framework for better performance. We systematically evaluate our method in three different segmentation tasks including CT kidney tumor, MR prostate, and PROMISE12 challenge, showing promising results compared with state-of-the-art methods. The code is available here: {\color{red}\url{https://github.com/sunalbert/Sequential-patch-based-segmentation}}.
\keywords{Medical Image Segmentation, Interactive Segmentation, Sequential Patches, ConvRNN}
\end{abstract}

\section{Introduction}
Accurate segmentation of different objects from medical imaging data (\emph{e.g.,} MR, CT) is normally believed as one of the most significant steps for clinical treatment. However, traditional segmentation is often performed manually by the physician. Unfortunately, this manual segmentation is extremely time-consuming, since currently large amount of data is being collected every day. Also, the inaccurate manual segmentation could not be totally avoided due to the factors like fatigue. Moreover, different physicians sometimes provide different segmentation according to her/his own experience. Thus, automatic or semi-automatic segmentation methods to quickly and precisely obtain the object boundary are urgently in demand.

As the deep learning goes popular in image segmentation task \cite{long2015fully} \cite{zhao2017pyramid} \cite{chen2016deeplab}, lots of releated attempts have been conducted for medical image segmentation. Ciresan \emph{et al. } \cite{ciresan2012deep} trained a patch-based network to predict the class label of the central point in each patch and won the EM segmentation challenge at ISBI 2012 by a large margin. Havaei \emph{et al.} \cite{havaei2017brain} used a two-pathway deep CNN to exploit both local features and global contextual features to improve segmentation. To achieve good localization and the use of context, Ronneberger \emph{et al.} \cite{ronneberger2015u} used skip connection to recover details lost by deep layers, which was generalized to other medical image segmentation tasks \cite{cciccek20163d} \cite{milletari2016v} \cite{yu2017volumetric} with promising results. 
%\vspace{-15pt}
\begin{figure}
\begin{center}
\includegraphics[width=1.0\linewidth]{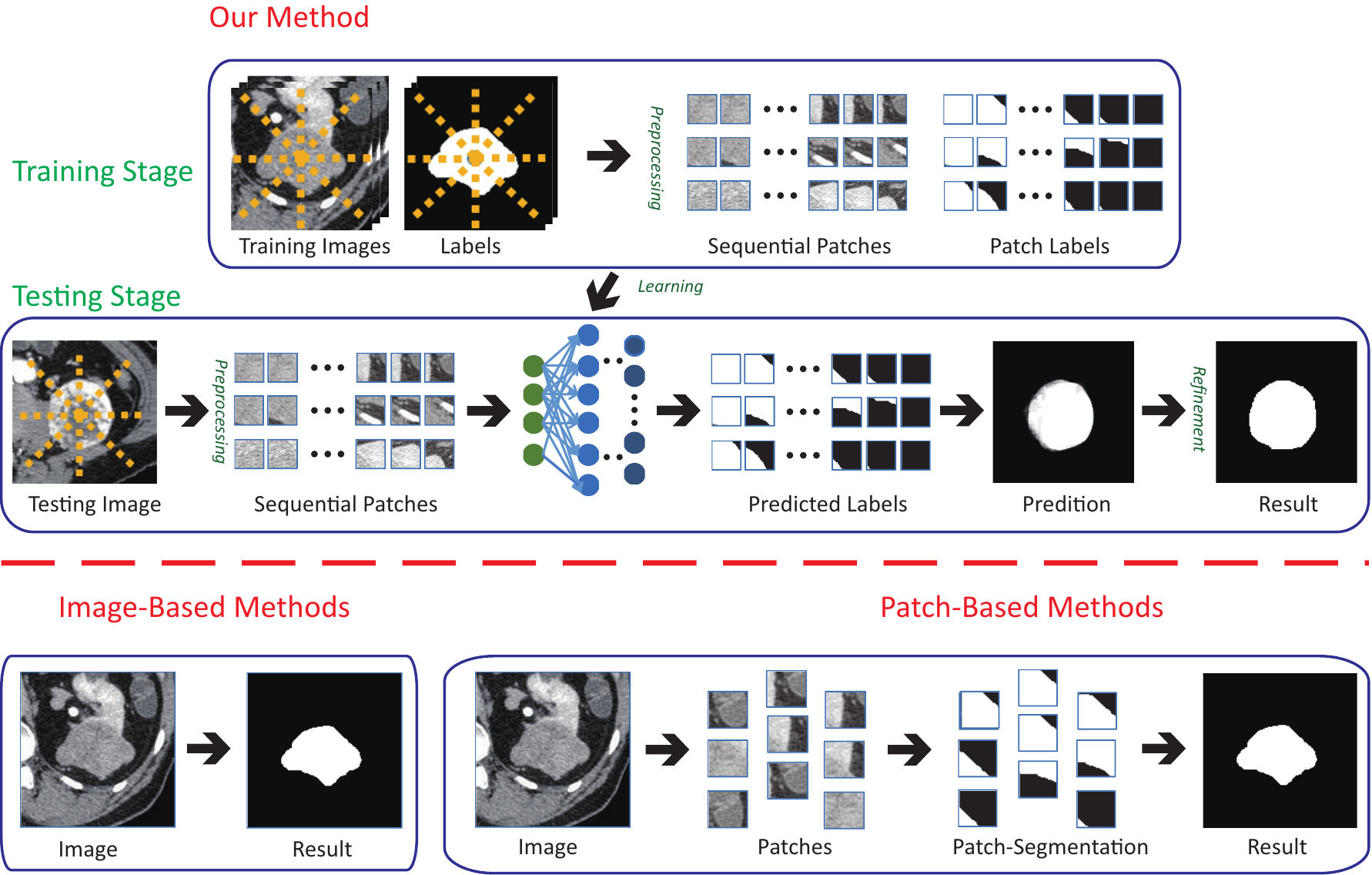}
\end{center}
   \caption{Flowchart of our method for interactive segmentation and the difference of our method from previous methods. Our method extracts sequential patches with 4 or 6 strides along 16 rays extending from the inside of object to outside (for better visualization, only 8 rays and sparsely sampled patches are shown). A sequence of patches along a certain ray is viewed as a single training sample. Image-based method takes the whole image as input and outputs its corresponding prediction directly. Patch-based method takes each patch as an individual sample.}
\label{fig:pipeline_comparison}
\end{figure}

Different from these deep learning methods, during manual delineation, the physician first locates the object roughly and keeps the appearance in mind, then repeatedly checks the inside-outside intensity changing to delineate the boundary precisely especially on the cases with blurry boundary. This reveals that the inside-outside comparison is important for medical image segmentation. Unfortunately, existing deep learning based methods directly classify each pixel/voxel after several convolution and pooling operations, which usually leads to inferior segmentation result on the cases with low contrast and blurry boundary.

In this paper, inspired by above observation, we propose a novel interactive segmentation method for medical images, by incorporating point-based interaction and bi-directional sequential patch learning. In our method, the physician is first asked to take a few seconds to click on the rough centeral point of the object before segmentation (we name the point as central point in this paper). With this click, an approximate location of object will be initially determined. As observed in Fig. \ref{fig:pipeline_comparison}, we can draw many rays (based on a rough central point given by the physician), extending from the inside of object to the outside. We extract patches with 4 or 6 strides along these rays resulting in several sequences of patches (we name as sequential patches in this paper). Normally, in a certain sequence of patches extracted along a ray, with a very high probability, the first several patches along the rays are inside the object and the last several patches are outside the object. Also, the patches along the direction of the rays change sequentially. Thus, we learn the feature representation of these sequential patches, to capture the shape and appearance changing from both the object-to-background and background-to-object directions, which is largely different from the previous pure patch-based and image-based segmentation methods (as shown in Fig. \ref{fig:pipeline_comparison}). Following this idea, we innovatively model the segmentation task as learning the representation of the bi-directional sequential patches, according to the given central point. To be specific, we build a U-net \cite{ronneberger2015u} like model with only 6 convolutions and 2 poolings. With the help of designed bi-directional gated ConvRNN modules embedded in different feature levels, our model takes sequential patches as input, thus the receptive field of all the layers (even the shallower layers) could be increased. In other words, we increase the receptive field via making the network goes wide, instead of deep. Besides, the bi-directional gated ConvRNN enables our model to capture spatial relation among adjacent patches and exploit more crucial details via inside-outside comparison as what physician does during segmentation. It is worth noting that the ConvRNN module in deeper layers could capture the spatial relation among patches in a general view, while the ConvRNN units in shallower levels will encode the detailed information for accurate segmentation. In brief, our work makes following contributions:

$\bullet$ We propose a simple point-based interactive method to improve the performance and meanwhile reduce the segmentation time. This point-based interaction could be naturally integrated with state-of-the-art deep learning method.

$\bullet$ Compared to traditional deep learning segmentation methods, we introduce inside-outside comparison into network, which has not been touched by previous works.

$\bullet$ We design a simple yet effective ConvRNN module. Thanks to this module, our model regards a sequence of patches as a single training sample and captures spatial relation among adjacent patches, which is different from patch-based methods (as shown in Fig. \ref{fig:pipeline_comparison}). Besides, the ConvRNN module could help the shallow network increase receptive field to capture more context.

$\bullet$ Our model is light (5.2MB), easy to train (3 hours on PROMISE12 dataset via Nvidia 1080), fast to test (80ms/slice on PROMISE12 testing dataset).

\section{Related Work}
Compared with natural image segmentation, medical image segmentation is required to face more challenges, \emph{e.g.,} low tissue contrast, large appearance variation, unpredictable object motion, \emph{etc}. Thus, traditional methods, specially developed for natural image segmentation, could not be directly applied to segment the medical objects. Recent years, many works have been contributed to medical image segmentation, which can be roughly classified into two categories.

The first class is registration/deformable model-based methods. Yang \emph{et al}. \cite{yang2014spatiotemporal} integrated the spatiotemporal information into a traditional registration-based method to perform effective 4-D MRI thoracic segmentation. Liao and Shen \cite{liao2012feature} employed both patient-specific information and population by using anatomical feature selection and an online updating mechanism. In \cite{zhuang2010registration}, locally affine registration method (LARM) was proposed to provide the correspondence of anatomical substructures, and the free-form deformations with adaptive control point status was performed to refine local details. In \cite{mesejo2016survey}, Chen \emph{et al}. imposed the anatomical constraints during the model deformation procedure. Gao \emph{et al.}  \cite{gao2016accurate} proposed to learn a displacement regressor which can provide a non-local external force for each vertex of deformable model.

The second class is learning-based methods, which have received the considerable attention in recent years, including the patch-based and image-based methods. Patch-based methods usually regard each individual patch as the input and predict its label as the output. For example, in \cite{li2015automatic}, Li \emph{et al.} designed a CNN to predict the label of input patch and combined the patch-level results to generate the final segmentation of liver tumor. Zhao \emph{et al.} \cite{zhao2016multiscale} integrated local and global region features into consideration and thus proposed a multi-scale convolution network to segment the brain tumor. To utilize the global context, as the image-based methods, whole image based deep models have also been explored. U-net \cite{ronneberger2015u} is one of the representative image-based methods. Several variants (\cite{clark2017fully} \cite{yu2017volumetric} \emph{etc}.) have been also proposed to segment different objects in medical images. Our method borrows the advantages from both image-based and patch-based methods, by modelling the sequential patches as aforementioned..

Moreover, from the perspective of interactive segmentation, our method belongs to the point-based interaction. Point-based interactive segmentation models have aroused increasing interest very recently. However, how to effectively integrate it with the state-of-the-art deep learning methods is still in its early stage. In \cite{bearman2016s}, the points given by human were regarded as weak supervision label for segmentation. Sun \emph{et al.} \cite{sun2017point} focused on generating prior map according to the given point to improve segmentation performance. Furthermore, the RNN has been introduced to image segmentation in recent years. Liang \emph{et al.} \cite{liang2016semantic} separated natural image with clear boundaries into different disjoint superpixels according to raw RGB pixels, then employed RNN to capture relations among superpixels. Cai \emph{et al. }\cite{cai2017improving} used RNN to capture the relations among adjacent slices. We claim that our method is fundamentally different from these previous ones:  the point in our method is just an indicator of the object; We capture the inside-outside changing trends with overlapped patches (which may contain both foreground and background) to determine the blurry boundary. Besides, our ConvRNN could be embedded into different levels of convnet instead of only the last few layers.

\section{Framework and Preprocessing}
We show the whole framework of our method in Fig. \ref{fig:pipeline_comparison}, including the training and testing stages. The training stage consists of the preprocessing and sequential patch learning, while the testing stage contains the preprocessing, sequential patch segmentation, and result refinement.

In the training stage, the preprocessing step aims to extract the sequential patches for the following learning step. Since the ground truth is available during the training stage, for each image, we first calculate its mass center (of the object) as the initialized point, and then extract the sequential patches along the rays extending from this point, which could fully capture the relation of continuous changing from the inside to outside. Afterwards, these obtained sequential patches with their respective labels will be employed to train our proposed model, which will be elaborated in Section \ref{sec_seq_learning}.

In the testing stage, to segment a new coming testing image, at first, an initial central point in the testing image will be roughly clicked by the physician, and then the sequential patches (according to the central point) will be extracted. Afterwards, these extracted sequential patches will be fed to the trained model to obtain the corresponding likelihood maps. Finally, these likelihood maps will be combined together for final segmentation in result refinement step.

\textbf{Preprocessing: }As shown in Fig. \ref{fig:pipeline_comparison}, we first extend rays (16 rays in experiment setting) from the initial point. For each ray, we then extract a sequence of patches with their size as $32 \times 32$. Since the spatial smoothness among these patches is important in our method, we indeed extract patches with 4 or 6 strides to guarantee that consecutive patches contain an overlap.  Since we found that the first several patches in a sequence are normally inside the object while the last several patches are normally outside the object, the boundary of object could appear in the rest several patches in the middle. Therefore, the advantage of our patch extraction strategy is that these patches could globally cover the whole object and also reduce the redundant background regions at the same time.

\section{Sequential Patch Learning}\label{sec_seq_learning}
\subsection{Problem Analysis}
With obtained sequential patches in hand, we can redefine the sequential patch-based segmentation as the problem of spatial sequence labeling. Normally, we denote a sequence of patches as $\textbf{X}$ and thus $\textbf{X}_t$ indicates the $t$-th patch in this sequence. The segmentation model takes patch $\textbf{X}_t$ as input at $t$-th step and outputs the corresponding segmentation result $\textbf{Y}_t$ based on its memory over previous patch segmentation and current input patch. The spatial relationship between adjacent patches are encoded in the memory of the segmentation model. Accordingly, the sequential patch-based segmentation can be formulated as a task to train a sequence labeling algorithm $\mathcal{H} : \textbf{X} \mapsto \textbf{Y}$, which can assign the most likely label to each patch in the length-$K$ input sequence:
\begin{align}
\widehat{\textbf{Y}}_{1}, \widehat{\textbf{Y}}_{2}, \cdots, \widehat{\textbf{Y}}_{K}  =  \mathop{\arg\max}_{\textbf{Y}_{1}, \textbf{Y}_{1}, \cdots, \textbf{Y}_{K}} p(\textbf{Y}_{1}, \textbf{Y}_{2}, \cdots, \textbf{Y}_{K} | \textbf{X}_{1}, \textbf{X}_{2}, \cdots, \textbf{X}_{K})
\end{align}
 
\begin{figure}[t]
\begin{minipage}{0.48\textwidth}
\includegraphics[width=\textwidth]{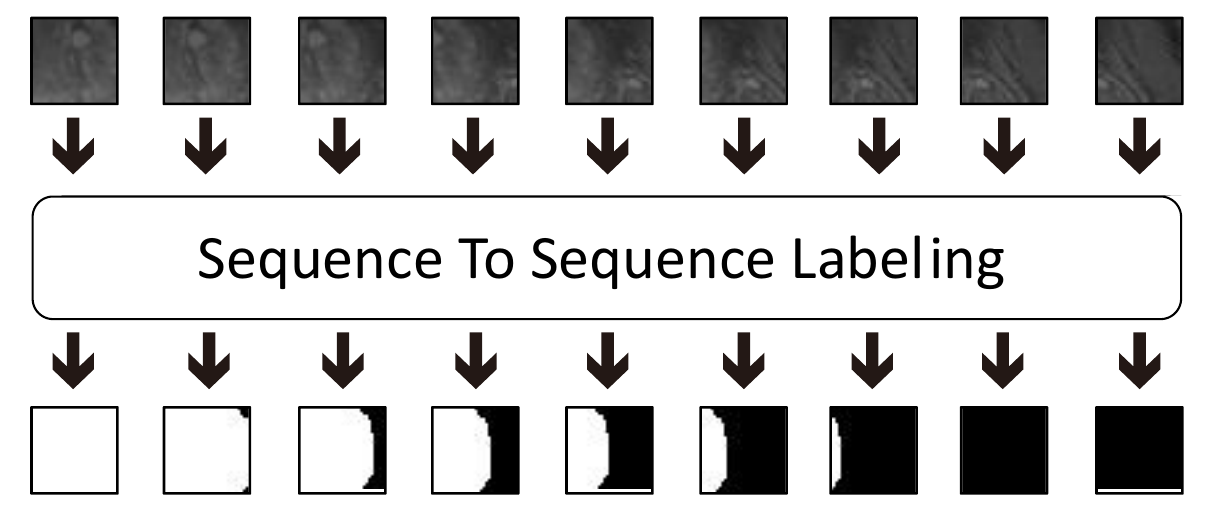}
\caption{Illustration of sequential patch-based segmentation problem. The input is a sequence of patches extracted from MR prostate and the output is the corresponding predictions.}
\label{fig:seq2seq}
\end{minipage}
\begin{minipage}{0.48\textwidth}
\includegraphics[width=\textwidth]{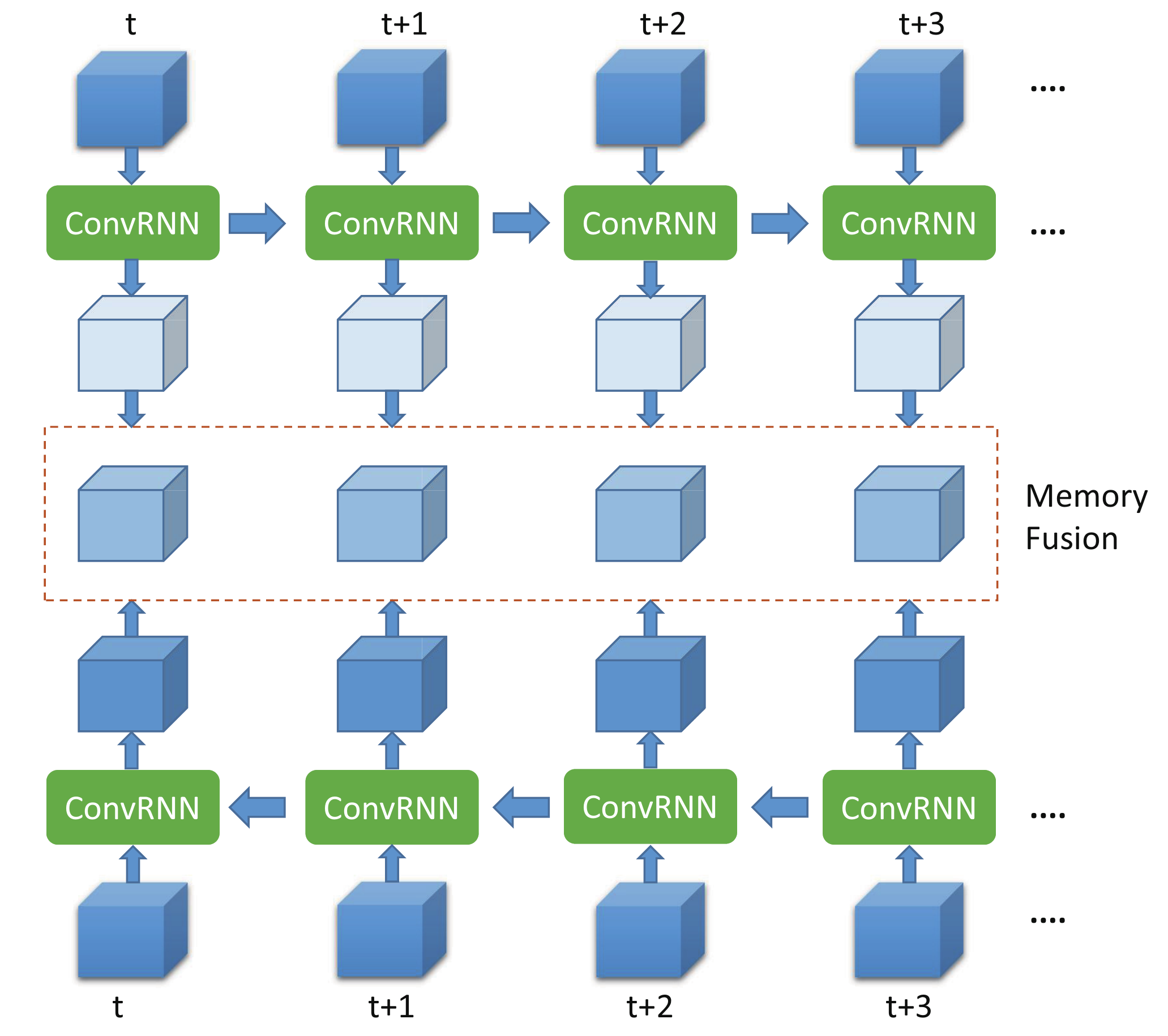}
\caption{Illustration of Bi-ConvRNN.}
\label{fig:bi_convrnn}
\end{minipage}
\end{figure}

It is worth noting that there is a key difference between our sequential patch-based segmentation and the traditional sequence labeling (in natural language process \cite{nguyen2007comparisons} \cite{ma2016end}): as shown in Fig. \ref{fig:seq2seq}, the output of each patch in a sequence is a 2D matrix instead of a scalar (in traditional sequence labeling). For an input patch with the size of $32 \times 32$, the number of its possible output pixels can be up to $2^{32}$. Therefore, it is difficult for existing models to well deal with the learning problem due to this large output volume. Thanks to our setting, with a high probability, the first several patches along the rays are inside the object and the last several patches are outside the object. Thus, our model actually more focuses on propagating label similarity from both ends to the middle patches, which could reduce the large output dimensionality and hence make the model more trainable.

\subsection{Gated Memory Propagation Unit}
Regarding that our goal here has been formulated as a sequence labeling task, it is a natural way to integrate this sequence labeling task into a recurrent neural network (RNN). As aforementioned, our patch extraction could guarantee the start (or end) patches to be foreground (or background), which could be treated as an important knowledge for learning.

As analyzed, our problem is to propagate the label with high confidence from both the ends (\emph{i.e.,} foreground or background) to the middle patches whose corresponding labels are with low confidence. To tackle this problem, we propose to use ConvRNN cell to build a gated propagation unit, which could capture the spatial relationship and propagate visual memory from current patch to its next patch. In particular, given a sequence of patches as the input, the propagation unit needs to encode the input into representative features and meanwhile propagate the label memory from the current patch to the next patch. Regarding that the boundary of medical object often shows low contrast, the propagation should be sensitive to slight intensity changing.

Therefore, we propose a simple yet effective ConvRNN cell as follows:
\begin{align}
\textbf{r}_t &= \sigma(\textbf{W}_{xr} \ast \textbf{X}_t + \textbf{W}_{hr} \ast \textbf{H}_{t-1} + b_r) \\
\textbf{H}_t &= relu(\textbf{W}_{xh} \ast \textbf{X}_t + \textbf{r}_t \circ \textbf{H}_{t-1} \ast \textbf{W}_{h\tilde{h}} + b_h)
\end{align}
where $\sigma$ is the sigmoid activation function, $\ast$, $\circ$ indicate the convolutional operator and element-wise multiplication, respectively. All the input $\textbf{X}_{1}, \cdots, \textbf{X}_{t}$, hidden state $\textbf{H}_{1}, \cdots, \textbf{H}_{t}$, and reset gate $\textbf{r}_{t}$ are 3D tensors in $\mathbb{R}^{P \times M \times N}$, where $P$ is the number of channels, $M$ is the number of rows, $N$ is the number of columns in feature map, respectively. The most important tensor is hidden state $\textbf{H}_t$ at $t$-th step, which encodes spatial relation among adjacent patches. The reset gate $\textbf{r}_t$ measures the correlation between current input and hidden state $\textbf{H}_{t-1}$, aiming to control that how much information from $\textbf{H}_{t-1}$ is allowed to flow into current hidden state $\textbf{H}_t$. If $\textbf{r}_t$ is close to zero, the memory from previous hidden state $\textbf{H}_{t-1}$ will be forgotten and meanwhile the new hidden state $\textbf{H}_t$ will be determined only by current input $\textbf{X}_t$ to handle the sudden change of intensity. The $relu$ \cite{nair2010rectified} activation function is employed to generate the final hidden state instead of $tanh$, since $tanh$ is easy to saturate and change the range of value in feature map. The hidden state is used directly as the new representation of $\textbf{X}_t$.

Given an input sequence starting from foreground (object) region and ending at background, a single direction with memory unit could only propagate the memory in forward or backward view. However, it is still hard to determine the location of stopping propagation for predicting the boundary of object. To improve the performance of our visual memory model, we innovatively employ bi-directional RNN for sequential patch learning to integrate the information from different two directions. As illustrated in Fig. \ref{fig:bi_convrnn}, the first unit processes the patches of a sequence in a forward direction, and the second unit processes the patches of a same sequence in a reversed direction. It is notable that the two units show different favors when propagating visual memory: forward unit tends to believe that current patch to be inside the object (classified as foreground), while the backward unit prefers to assign negative label (classified as background) to current patch. Finally, we employ a convolution layer with $1 \times 1$ kernel as a memory fusion module to fuse the different memories.

\subsection{Network Structure}
\begin{figure}
\begin{center}
\includegraphics[width=0.95\linewidth]{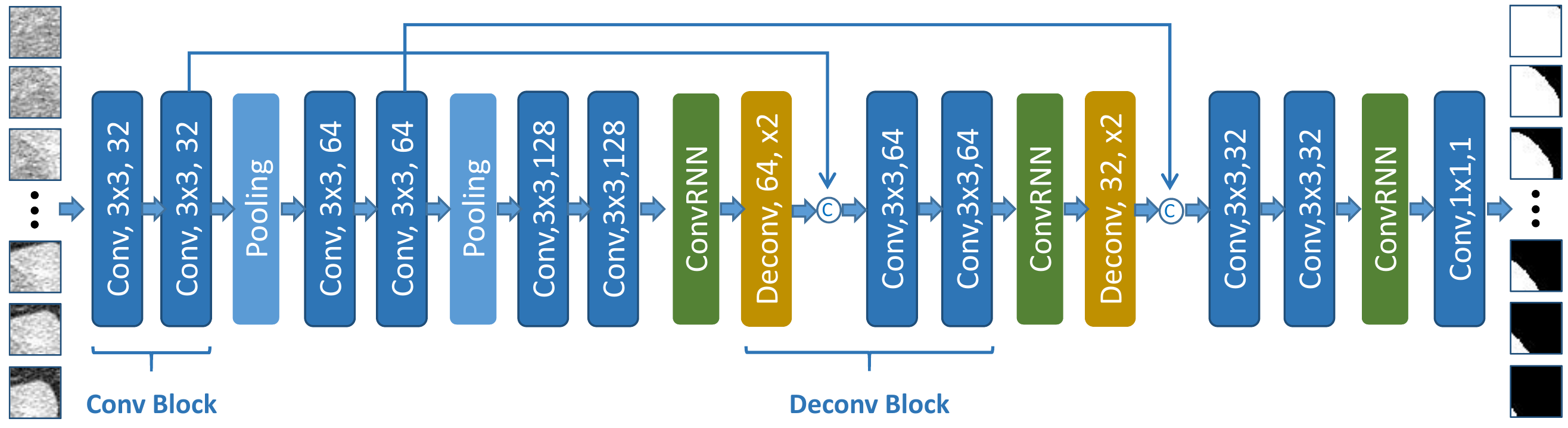}
\end{center}
   \caption{Details of our model. The high resolution feature maps from contracting path are concatenated with the output of deconvolution operation.}
\label{fig:netstructure}
\end{figure}

Inspired by vanilla U-net \cite{ronneberger2015u}, our network also contains a contracting path and an expanding path (see Fig. \ref{fig:netstructure}). The contracting path consists of three conv blocks, in which each block contains two time-distributed convolution layers. The max pooling operation between two contracting blocks aims to keep the notable features and increase the size of receptive field of network. The two deconv blocks in expanding path are employed to increase the size of feature map.

Compared with common ways of using ConvGRU \cite{tokmakov2017learning} and ConvLSTM \cite{xingjian2015convolutional}, we embed the ConvRNN modules into the different levels of our model. Our consideration includes that (1) the features from the deeper layers are more effective to measure the correlation between two patches in a general view, because the size of receptive field will increase when the convolution and pooling operations are employed; (2) the features from the shallower layers contain more details which is useful to delineate the boundary of object precisely (3) from another perspective, the ConvRNN embedded in the top level (where the feature maps share the same size with origianl patch) performs like learning pixel-wise non-linear affinity matrix which proves to be effective for segmentation refinement \cite{liu2017learning} \cite{maire2016affinity}. As illustrated in Fig.\ref{fig:netstructure}, the feature maps are fed into a gated memory propagation module aiming to learn spatial relationship between adjacent patches. Then, the output of ConvRNN are upsampled and concatenated with the feature map from the contracting path. In other words, the detailed spatial relation at certain level is captured by both memory generated in a general view and the feature map from corresponding level in the contracting path.

Since every aforementioned component is differentiable, our model can be trained in an end-to-end way. We employ the binary cross entropy as loss function in our model. We apply Adam \cite{kingma2014adam} with $\beta_1=0.95, \beta_2=0.99$ to optimize our model. Besides, weight decay is employed to avoid overfitting. We combine the patch-level segmentation result to whole slice to generate the final segmentation result in refinement.

\section{Results}
We report the qualitative and quantitative segmentation results of our method on three medical image segmentation datasets, including CT kidney tumor segmentation, MR prostate segmentation, and PROMISE12 challenge. For the evaluation metrics, we employ the Dice Ratio Score (DSC) ($\%$) and Centroid Distance (CD) (mm) along 3 different directions (\emph{i.e.,} lateral x-axis, anterior-posterior y-axis, and superior-inferior z-axis), which are widely used in previous literatures\cite{shi2015semi} \cite{shi2017does}. Also, for PROMISE12 challenge, several additional metrics (\emph{e.g.,} ABD, 95HD) are reported according to the automatic calculation provided by the challenge organizers \cite{litjens2014evaluation}.

\subsection{CT Kidney Tumor Segmentation}
\textbf{Setting:} Our CT kidney dataset consists of 60 patients (images) with about 2500 CT slices, scanned from 60 different patients. The resolution of each CT image after image preprocessing is $296 \times 296 \times 40$, with the in-plane voxel size as $1 \times 1$ mm$^2$ and the inter-slice thickness as $1$ mm. The manual delineation results are available for all the images which could be considered as the ground truth for performance evaluation. For each input sequential, we extract 15 patches with 4 strides. For segmenting a new coming testing image, we will first ask the physician to perform one click to indicate the rough center of the tumor. Basically, we perform 2-fold cross validation on these 60 CT images.

\begin{figure}[t]
\begin{center}
	\includegraphics[width=0.9\linewidth]{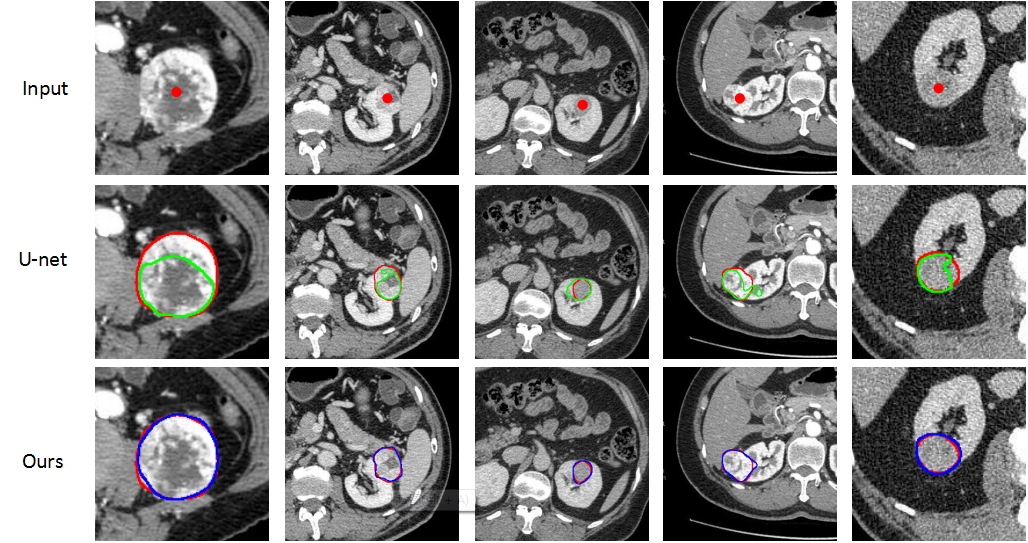}
\end{center}
   \caption{Typical result of tumor segmentation. Slices in the first row are the input for network. The red dots given by physicians indicates the rough center area of tumor. The red, green and blue curves denote the ground truth, result of U-net and our method, respectively.}
\label{fig:method_comparision}
\end{figure}

\begin{figure}
\subfigure[] { \label{fig:vis:top_level} 
\includegraphics[width=0.5 \textwidth]{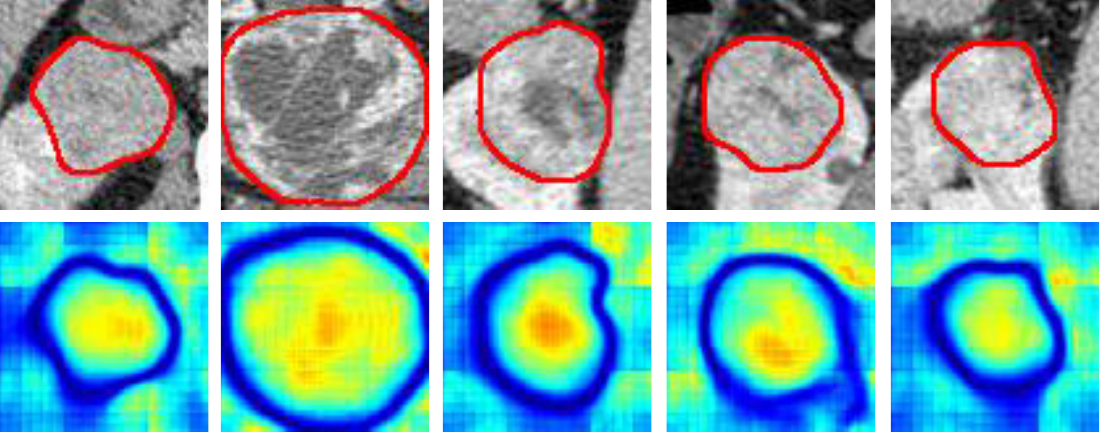}
}
\subfigure[] { \label{fig:vis:multi_level} 
\includegraphics[width=0.5 \textwidth]{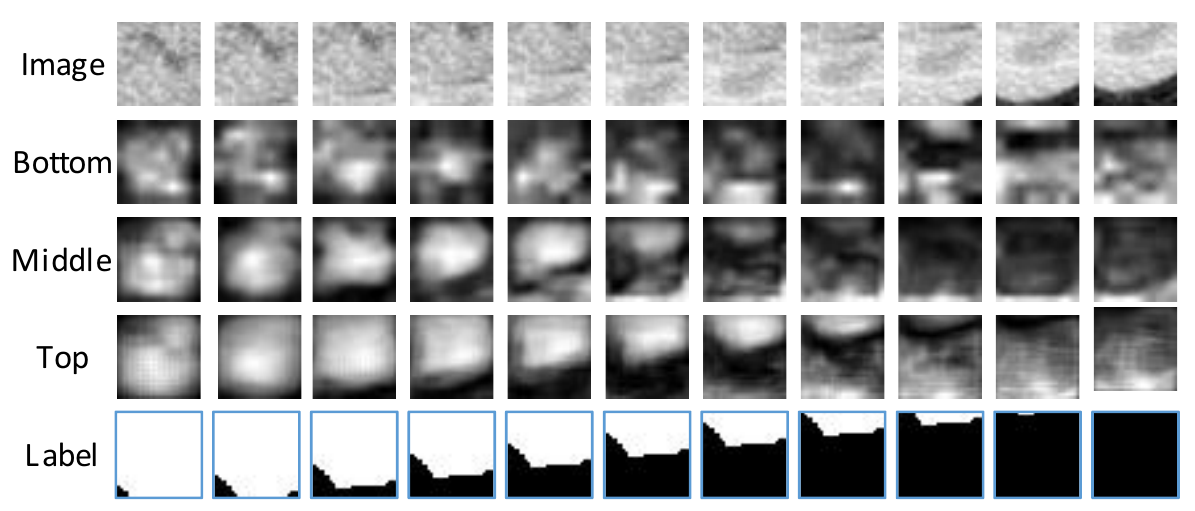}
}
\caption{The visualization of our method. As expected, our method shows the ability to distinguish the object from the background by a large margin. (a) The visualization of fused top level feature maps. (b) The visualization of multi-level feature maps.}
\label{fig:vis}
\end{figure}

\begin{table}[t]
\small
\begin{center}
\begin{tabular}{lcccc}
\hline
Method & Patch-based FCN & U-net & Ours\\
\hline
CD-x(mm) & - & $0.71 \pm 0.52$ &  \textbf{0.57 $\pm$ 0.26}\\
CD-y(mm) & - &  $0.59 \pm 0.64$ &  \textbf{0.33 $\pm$ 0.26}\\
CD-z(mm) & - &  $0.37 \pm 0.32$ & \textbf{0.25 $\pm$ 0.14}\\
DSC[\%] & 34.45 & $86.52 \pm 9.12$ &  \textbf{93.24 $\pm$ 4.12}\\
\hline
\end{tabular}
\end{center}
\caption{Comparison with other methods on CT kidney tumor dataset. The result of patch-based FCN is very inferior, thus we only present the DSC. Our method outperforms other methods.}
\label{tab:tumor_comparison}
\end{table}

\textbf{Comparison with Baselines:} We compare our model with two baselines, \emph{e.g.,} U-net (image-based method) \cite{ronneberger2015u} and patch-based FCN. It is worth noting that, for fair comparison, the U-net in our experiment is designed with enough receptive field to cover common kidney tumors. For the implementation of patch-based FCN, we extract 1000 patches for each slice as the following way: (1) densely sampling the patches that are located inside the tumor or  centered close to the tumor boundary, and (2) sparsely sampling the patches that are far from the tumor boundary. Table \ref{tab:tumor_comparison} reports the quantitative result. We can observe that the patch-based FCN performs worst, which is because that treating image patches individually as the direct input of the model might ignore the context information of the whole CT slice. For patch-based FCN, it is difficult to distinguish two patches (inside and outside the tumor) with similar appearance according to its setting. Moreover, U-net \cite{ronneberger2015u} performs better than patch-based FCN. As shown in Fig. \ref{fig:method_comparision}, U-net is able to roughly locate the tumor by using the whole image context information. However, it still fails to delineate the tumor precisely when the texture of tumor is similar to that of the renal parenchyma. Normally, the skip connection in U-net is applied to refine the roughly segmentation generated by deeper layers. Unfortunately, there is a shortcoming of this approach: the receptive field of the shallower layers is small and limited (usually $7 \times 7$ or $15 \times 15$), thus the learned feature maps are not discriminative enough to determine the boundary in low contrast region. Different from U-net, all the layers in our network keep large receptive field (even the shallower layers) and exploit crucial inside-outside changing trends as what physician does during delineation. It is worth noting that our model is much lighter (5.2MB) than U-net (135.9MB). We illustrate several typical segmentation examples in Fig. \ref{fig:method_comparision}, showing the promising performance compared with U-net. We also show the visualization of the feature maps of our method in Fig. \ref{fig:vis}. With the help of inside-outside comparison, our method shows the ability to be aware of the blurry boundary.

\begin{figure}[t]
\begin{minipage}[t]{0.48\textwidth}
\includegraphics[width=\textwidth]{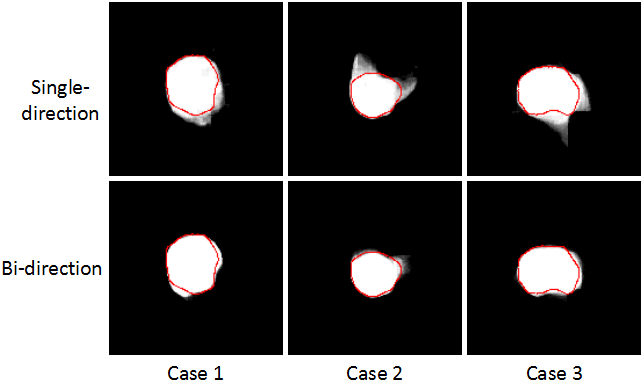}
\caption{Tumor-likelihood maps generated by single directional arechitecture and bi-directional arechitecture.}
\label{fig:single_vs_bi}
\end{minipage}
\begin{minipage}[t]{0.48\textwidth}
\includegraphics[width=\textwidth]{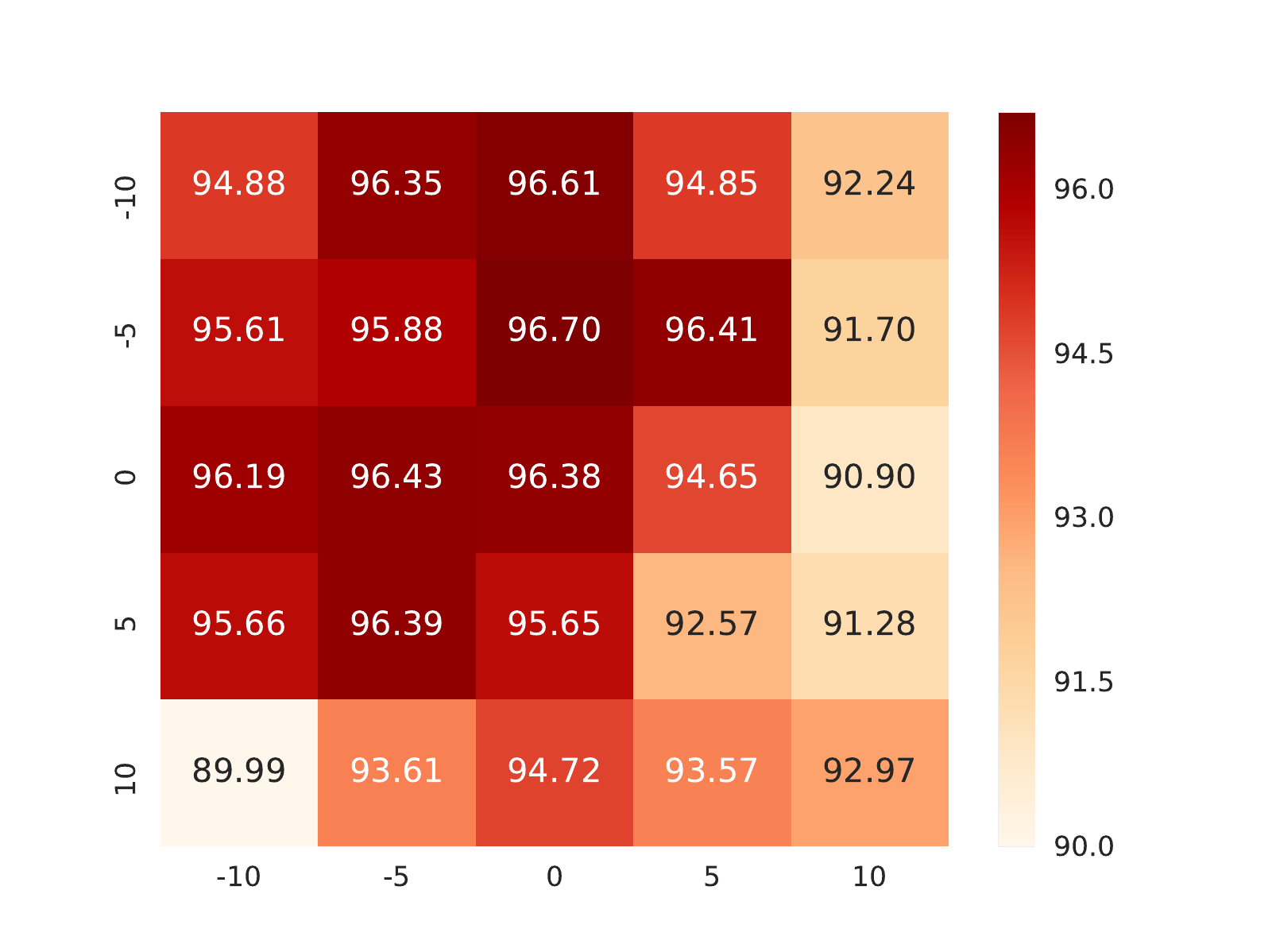}
\caption{Illustration of DSC with different initial coordinates.}
\label{fig:heatmap}
\end{minipage}
\end{figure}

\textbf{Ablation Study:}
To systematically study the efficacy of each component in our method, several ablation experiments are conducted. Table \ref{tab:ablation_study} reports the corresponding results. Typically, Vanilla U-net without ConvRNN performs much
\begin{table}
\begin{center}
\begin{tabular}{cccccccc}
\hline
Method & U-net & Seq-FCN & U-net-B & U-net-LSTM & Single direction & U-net-BM & Ours  \\
\hline
DSC[\%] & 86.52 & 90.42 & 91.28 & 91.51 & 92.31 & 92.34 & \textbf{93.24} \\
\hline
\end{tabular}
\end{center}
\caption{Ablation study results. Seq-FCN denotes the seq-based FCN. U-net-B, U-net-BM denote the U-net embedded with ConvRNN in the bottom level, both the bottom middle level, respectively.}
\label{tab:ablation_study}
\end{table}
worse than our method, with only $86.52\%$ on DSC. Also, we first embed a ConvRNN block into the bottom level of U-net and then feed sequential patches, leading to a $4.76\%$ improvement. The U-net with ConvRNN embedded into both the bottom and middle level further advances the DSC to $92.34\%$. As observed, the ConvRNN could successfully capture the spatial relation, incorporate the context to improve the performance. Moreover, we modify the patch-based FCN to seq-based FCN by embedding ConvRNN into its original structure. Compared to patch-based FCN, seq-based FCN achieves substantial improvement on DSC, which reveals the efficacy of our sequential patches. Furthermore, we evaluate the single direction design of our ConvRNN. As shown in Fig. \ref{fig:single_vs_bi}, without the spatial relation from the bi-direction and the memory fusion, this architecture leads to a slight drop in DSC. Finally, we replace our ConvRNN with ConvGRU \cite{tokmakov2017learning} or ConvLSTM \cite{xingjian2015convolutional} to evaluate the influence. Unfortunately, the network with ConvGRU \cite{tokmakov2017learning} is hard to train and fails to get a good performance. The variant with ConvLSTM \cite{xingjian2015convolutional} performs a little worse than our method especially on these tumors with blurry boundary. One possible reason for this problem is that the long time memory encoded in the cell state is not sensitive to slight intensity changing, thus the ConvLSTM\cite{xingjian2015convolutional} unit may get inferior result around the blurry boundary area.

\textbf{Influence of Initial Click Location:}
As an interactive segmentation method, to evaluate how different initial click locations (different coordinates) affect the segmentation result, we randomly choose the $10$-th slice of the $31$-st patient with different initial (one) points. As what Fig. \ref{fig:heatmap} conveys, the coordinates of the different initial points really influence the result.
Specifically, (0,0) indicates the ground truth of the central point, and different numerical values in x and y-axis mean different offsets according to the central point.
Basically, the farther away from the real central point of tumor, the worse the result will be. However, it is noteworthy that our method still can obtain the robust results and also outperform vanilla U-net on this slice. Also, it is a rare case for an experienced physician to click a very incorrect central point of the object.

\begin{figure}
\small
\begin{center}
	\includegraphics[width=1.0\linewidth]{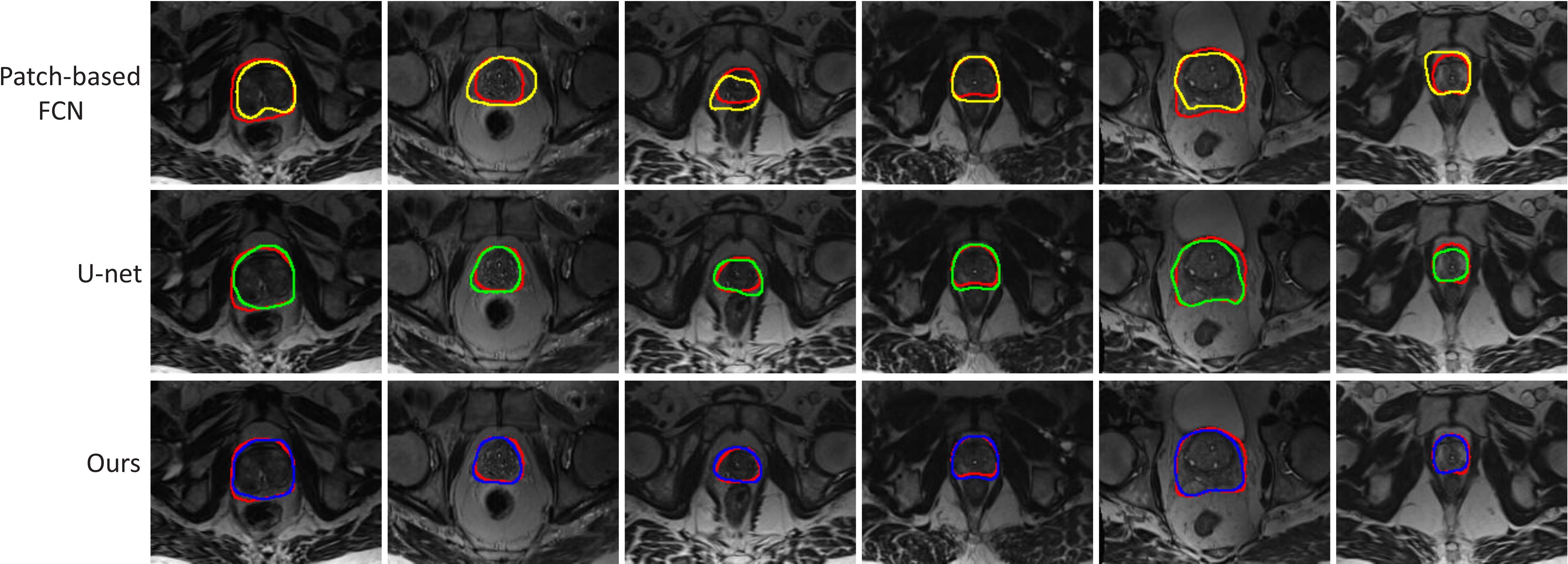}
\end{center}
   \caption{Typical results. The red curve denotes the ground truth. The yellow, green, blue curves denote the results of patch-based FCN, U-net and our method, respectively.}
\label{fig:mr_prostate_result}
\end{figure}

\subsection{MR Prostate Segmentation}
\textbf{Setting:}
We also validate our method on a  MR prostate segmentation dataset. This dataset is collected by scanning 22 different patients. The ground truth of prostate region is manually delineated by the physician for comparison. The resolution of MR images after preprocessing is $193 \times 153 \times 60$. We randomly select 11 patients as the training set and use the remaining patients as the testing set. All the parameters in our model keep the same as that in segmenting CT kidney tumor. Compared to the tumor, the size of prostate is relative smaller, thus, for each input sequential patches, we extract 8 patches with 4 strides.

\begin{table}
\begin{center}
\begin{tabular}{lcccc}
\hline
Method & Patch-based FCN & U-net \cite{ronneberger2015u} & Ours\\
\hline
CD-x(mm) & $1.91 \pm 2.63$ & $0.76 \pm 0.45$ & \textbf{0.47 $\pm$ 0.34}\\
CD-y(mm) & $1.31 \pm 0.99$ &  $0.56 \pm 0.64$ & \textbf{0.28 $\pm$ 0.16}\\
CD-z(mm) & $0.57 \pm 0.37$ &  $0.19 \pm 0.32$ & \textbf{0.12 $\pm$ 0.23}\\
DSC[\%] & $71.35 \pm 21.04$ & $84.69 \pm 13.21$ & \textbf{90.91 $\pm$ 6.32} \\
\hline
\end{tabular}
\end{center}
\caption{Comparison with other methods on MR prostate dataset. Our method achieves best performance.}
\label{tab:mr_prostate}
\end{table}

\textbf{Results:}
Fig. \ref{fig:mr_prostate_result} illustrated several typical results of our method on MR prostate segmentation. As observed, our method can precisely segment the prostate boundary compared with two baselines, even for these difficult cases (\emph{i.e.}, top and bottom slices). We also show the numerical results of our method, patch-based FCN and U-net \cite{ronneberger2015u} in Table \ref{tab:mr_prostate}, showing the best performance achieved by our method on both DSC and CD.

\subsection{Promise12 Prostate Segmentation Challenge}
\textbf{Setting:}
We also evaluate our method on MICCAI Prostate MR Image Segmentation (PROMISE12) challenge dataset, which is usually considered as the most famous challenges in MR prostate segmentation. In PROMISE12, the training dataset consists of 50 T2-weighted MR images of prostate with the corresponding ground truth. The organizers also provide additional 30 MR images as testing without the ground truth. Thus, after segmentation, we submitted the results of the testing images to the organizers, and the organizers calculated the quantitative results. Specifically, beyond DSC, the percentage of the absolute difference between the volumes (ABD), the average over the shortest distances between the boundary points of the volumes (aRVD) and the $95\%$ Haussdorf distance (95HD) are employed as the evaluation metrics. Since these MR images are collected from different medical institutions, there are large variance of voxel spacing among different MR images. 
\begin{figure}[t]
\begin{center}
  \includegraphics[width=1.0\linewidth]{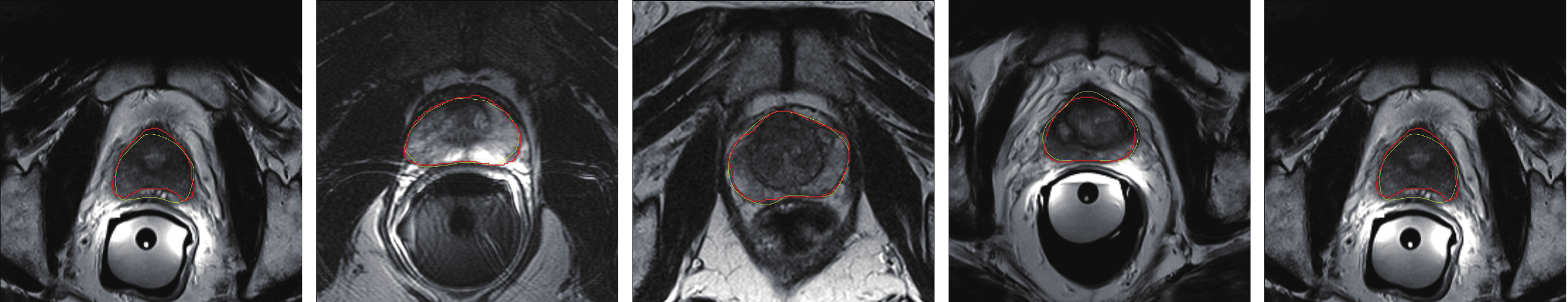}
\end{center}
   \caption{Several typical results of our method on testing dataset. The yellow and red curve denote the ground truth and our results, respectively.}
\label{fig:promise12_result}
\end{figure}
In order to reduce the influence of spacing, we keep the spacing in z-axis unchanged and resize these images into $0.625 \times 0.625 $ mm in x-axis and y-axis. As Fig. \ref{fig:pipeline_comparison} shows, we extract the sequential patches from 16 different rays extending from the inside of prostate to the outside with the stride as 6. Also, each sequence consists of 12 patches with size of $32 \times 32$.

\begin{table}[!htbp]\scriptsize
\centering
\begin{tabular}{l|ccc|ccc|ccc|ccc|c}
  \hline
  \multirow{2}{*}{Method} &\multicolumn{3}{c|}{ABD[mm]} &\multicolumn{3}{c|}{95HD[mm]} &\multicolumn{3}{c|}{DSC[\%]}  &\multicolumn{3}{c|}{aRVD[\%]} &\multirow{2}{*}{Score}\\
                                      &Whole        &Base         &Apex  &Whole        &Base        &Apex   &Whole        &Base      &Apex  &Whole        &Base       &Apex & \\
  \hline
    CAMP-TUM2	                &2.23         &2.46         &2.03  &5.71         &5.84        &4.62   &86.91        &84.31     &85.40  &14.98        &20.84      &21.21 &82.39 \\
    OncoB		           &2.20         &2.20         &2.47  &5.85         &\textbf{5.15}&5.65   &86.99       &86.09     &79.24  &13.35        &14.74      &28.65 &82.72 \\
    UdeM 2D			            &2.17         &2.39         &2.07  &6.12         &6.44        &4.71   &87.42        &84.93     &84.16  &12.37        &18.37      &21.90 &83.02 \\
    ScrAutoProstate               &2.13         &2.23         &2.18  &5.58         &5.60        &4.93   &87.45        &86.30     &83.47  &13.56        &14.46      &23.78 &83.49 \\
    Emory	             &2.14         &2.65         &2.41  &5.94         &5.45        &4.73   &87.99        &86.06     &84.53  &8.64        &15.70      &20.32 &83.66 \\
    Imorphics                       &2.10         &2.18         &1.96  &5.94         &5.45        &4.73   &87.99        &86.06     &84.53  &11.65        &13.33      &20.75 &84.36 \\
    methinks                        &2.06         &\textbf{2.11}&2.01  &5.53         &5.45        &4.62   &87.91        &86.79     &84.58  &8.71        &\textbf{10.84}      &21.21 &85.41 \\
    CREATIS                        &1.93         &2.14         &1.74  &5.59         &5.62    &\textbf{4.22}&89.33      &86.60     &86.77  &9.20        &14.65      &16.64 &85.74 \\
    CUMED			            &1.95         &2.13         &1.74  &5.54         &5.41        &4.29   &89.43        &86.42     &\textbf{86.81}  &\textbf{6.95}        &11.04      &\textbf{15.18} &86.65 \\
    \hline
    MedicalVision      &\textbf{1.79}&\textbf{2.11}&2.29  &\textbf{5.35}&6.20        &5.86 &\textbf{89.81}&\textbf{87.49}&81.74  &8.24        &11.52      &19.40 &85.33 \\
  \hline
\end{tabular}
\caption{Comparison with methods proposed by other competitors.}
\label{tab:promise12_comparison}
\end{table}

\textbf{ Results:}
Our method achieves 85.33 total score, leading to a fourth place (\emph{i.e.,} MedicalVision) on the leaderboard (till Jan.2018). Fig. \ref{fig:promise12_result} shows several typical results of our method on the testing dataset. As observed, our method can localize the prostate and delineate the boundary accurately.

The results of top 10 teams in PROMISE12 are shown in Table. \ref{tab:promise12_comparison}. Seven of ten teams utilize deep learning method (ScrAutoProstate, Emory and Imorphics are traditional methods). Except for UdeM2D \cite{drozdzal2017learning} and ours, remaining deep learning methods perform segmentation on 3D volume. Our method achieves competitive results compared to the state-of-the-art methods: CUMED \cite{yu2017volumetric}. Also, our method obtains the best performance on several metrics (\emph{i.e.,} Whole ABD, Base ABD, Whole 95HD, Whole DSC and Base DSC), which shows the effectiveness of our method.

Compared to the 3D methods, our method can be directly borrowed to deal with 2D medical image segmentation tasks. Also, our method has its own potential to extend to its 3D version by fusing the segmentation results from different directions. Moreover, please note that, for the point-based interaction, we click the central point according to our own experience. This will bring the noisy especially for the top and bottom slices (our Whole DSC and Base DSC are the best, while Apex DSC is worse than other methods).

\section{Conclusion}
In this paper, we present a novel method for medical image segmentation using sequential patches, by imitating inside-outside comparison as what physician does during manual delineation. During the segmentation, we first ask the physician to take a few seconds to click on the rough central point of the object. Then, according to this point, we extract different sequential patches and hence train a sequential patch learning model for prediction, by designing a specific gated memory propagation unit. Finally, as U-net, a multi-layer architecture is performed. Besides, our model light (5.2MB), easy to train and fast to test. We evaluate our method on CT kidney tumor segmentation, MR prostate segmentation, and PROMISE12 challenge, showing promising results.

\clearpage

\bibliographystyle{splncs03}
\bibliography{eccv2016submission}

\title{Supplementary materials} % Replace with your title
\author{}

%Please write out author names in full in the paper, i.e. full given and family names. 
%If any authors have names that can be parsed into FirstName LastName in multiple ways, please include the correct parsing, in a comment to the volume editors:
%\index{Lastnames, Firstnames}
%(Do not uncomment it, because you may introduce extra index items if you do that...)

\institute{
}

\maketitle

In this supplementary, we report the results of additional experiments.

\section{Additional Experiments}
We here consider the semi-automatic settings for patch-based FCN and patch-based U-net. Specially, we evaluate the patch-based FCN and patch-based U-net on patches which are extracted from the rays extending from the inside of the object to the outside. In other words, the patches are extracted in the same way as our method, while patch-based FCN and patch-based U-net still take a single patch as a training sample. Moreover, in the testing stage, the point in the center of the object is given by the physician. Then we fed the patches along the rays extending from the initial point to the outside of the object into patch-based FCN and patch-based U-net. 
\begin{figure}
\begin{center}
   \includegraphics[width=0.8\linewidth]{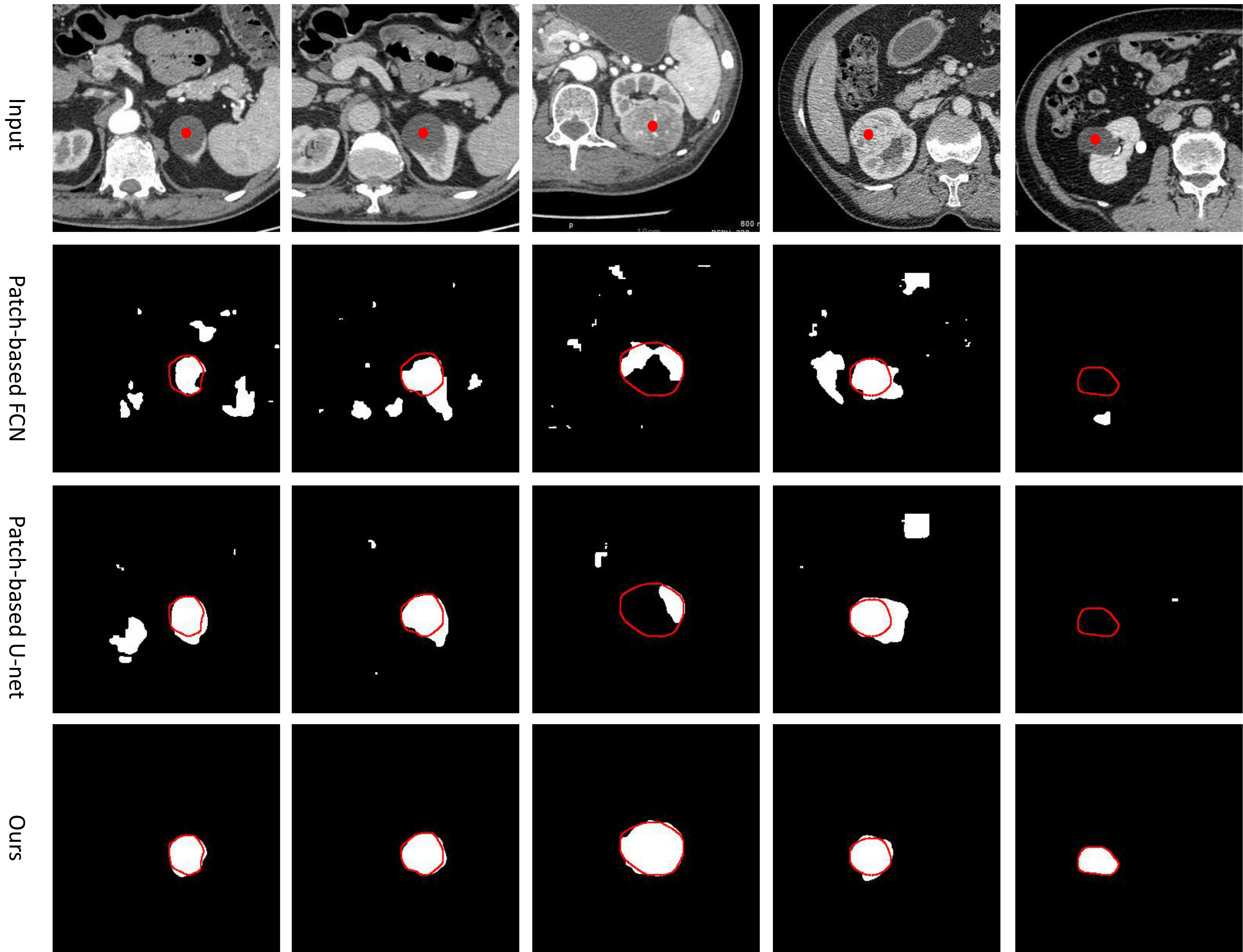}
\end{center}
   \caption{Typical results on kidney tumor dataset. The red dots are the central points given by the physician. The red curves denote the ground truth delineated by the physician.}
\label{fig:kidney_fcn}
\end{figure}

\begin{table}
\small
\begin{center}
\begin{tabular}{lcccc}
\hline
Method &  CD-x & CD-y & CD-z & DSC[\%] \\
\hline
Patch-based FCN &  $10.12 \pm 232.80$ & $9.97 \pm 132.93$ &  $8.23 \pm 60.16$ & $69.73 \pm 5.5$\\
Patch-based U-net & $9.07 \pm 193.18$ &  $15.19 \pm 763.64$ &  $7.47 \pm 55.84$ &  $75.55 \pm 8.1$\\
Ours &  $\textbf{0.57} \pm \textbf{0.26}$ & $\textbf{0.33} \pm \textbf{0.26}$ & $\textbf{0.25} \pm \textbf{0.14}$ & $\textbf{93.24} \pm \textbf{4.12}$\\
\hline
\end{tabular}
\end{center}
\caption{Comparison with patch-based FCN and patch-based U-net which are both trained on patches extracting along the rays extending from the inside of object to the outside. Our method with CRF performs best.}
\label{tab:tumor_comparison}
\end{table}

The quantitative results are shown in Table. \ref{tab:tumor_comparison}. As shown in Fig. \ref{fig:kidney_fcn}, although the central point is given by the physician, the patch-based FCN and patch-based U-net still fail to locate the tumor and delineate the boundary of the tumor precisely.

\end{document}